\def\year{2022}\relax
\documentclass[letterpaper]{article} 
\usepackage{aaai22}  
\usepackage{times}  
\usepackage{helvet}  
\usepackage{courier}  
\usepackage[hyphens]{url}  
\usepackage{graphicx} 
\usepackage{natbib}  
\usepackage{caption} 
\DeclareCaptionStyle{ruled}{labelfont=normalfont,labelsep=colon,strut=off} 
\usepackage{xcolor}
\usepackage{xspace}

\newcommand{\approach}{{LADC}\xspace}
\newcommand{\tool}{{LADC}\xspace}
\usepackage{multirow}
\usepackage{bm}
\usepackage{booktabs}
\usepackage{amssymb}
\usepackage{bbding}
\begin{document}
%

\title{Label-Aware Distribution Calibration for Long-tailed Classification}
\author{Chaozheng Wang\equalcontrib\textsuperscript{\rm 1},Shuzheng Gao\equalcontrib\textsuperscript{\rm 1},Pengyun Wang\textsuperscript{\rm 2}, Cuiyun Gao\textsuperscript{\rm 1}\thanks{Corresponding author},\\Wenjie Pei\textsuperscript{\rm 1},Lujia Pan\textsuperscript{\rm 2},Zenglin Xu\textsuperscript{\rm 1} }

\affiliations {
\textsuperscript{\rm 1} Harbin Institute of Technology, Shenzhen,  \\
\textsuperscript{\rm 2} Noah's Ark Lab, Shenzhen, Huawei Technologies Co. Ltd, \\

wangchaozheng@stu.hit.edu.cn,
\{szgao98,zenglin\}@gmail.com,
\{wangpengyun,panlujia\}@huawei.com,
gaocuiyun@hit.edu.cn,
wenjiecoder@outlook.com
}

\maketitle
\begin{abstract}
\begin{quote}

Real-world data usually present long-tailed distributions. Training on imbalanced data tends to render neural networks perform well on head classes while much worse on tail classes. The severe sparseness of training instances for the tail classes is the main challenge, which results in biased distribution estimation during training. Plenty of efforts have been devoted to ameliorating the challenge, including data re-sampling and synthesizing new training instances for tail classes. However, no prior research has exploited the transferable knowledge from head classes to tail classes for calibrating the distribution of tail classes. In this paper, we suppose that tail classes can be enriched by similar head classes and propose a novel distribution calibration approach named as \textbf{L}abel-\textbf{A}ware \textbf{D}istribution \textbf{C}alibration (\tool). \tool transfers the statistics from relevant head classes to infer the distribution of tail classes. Sampling from calibrated distribution further facilitates re-balancing the classifier. Experiments on both image and text long-tailed datasets demonstrate that \tool significantly outperforms existing methods. The visualization also shows that \tool provides a more accurate distribution estimation.

\end{quote}
\end{abstract}

\section{Introduction}

\begin{figure}[t]
    \centering
    \includegraphics[width=0.48\textwidth]{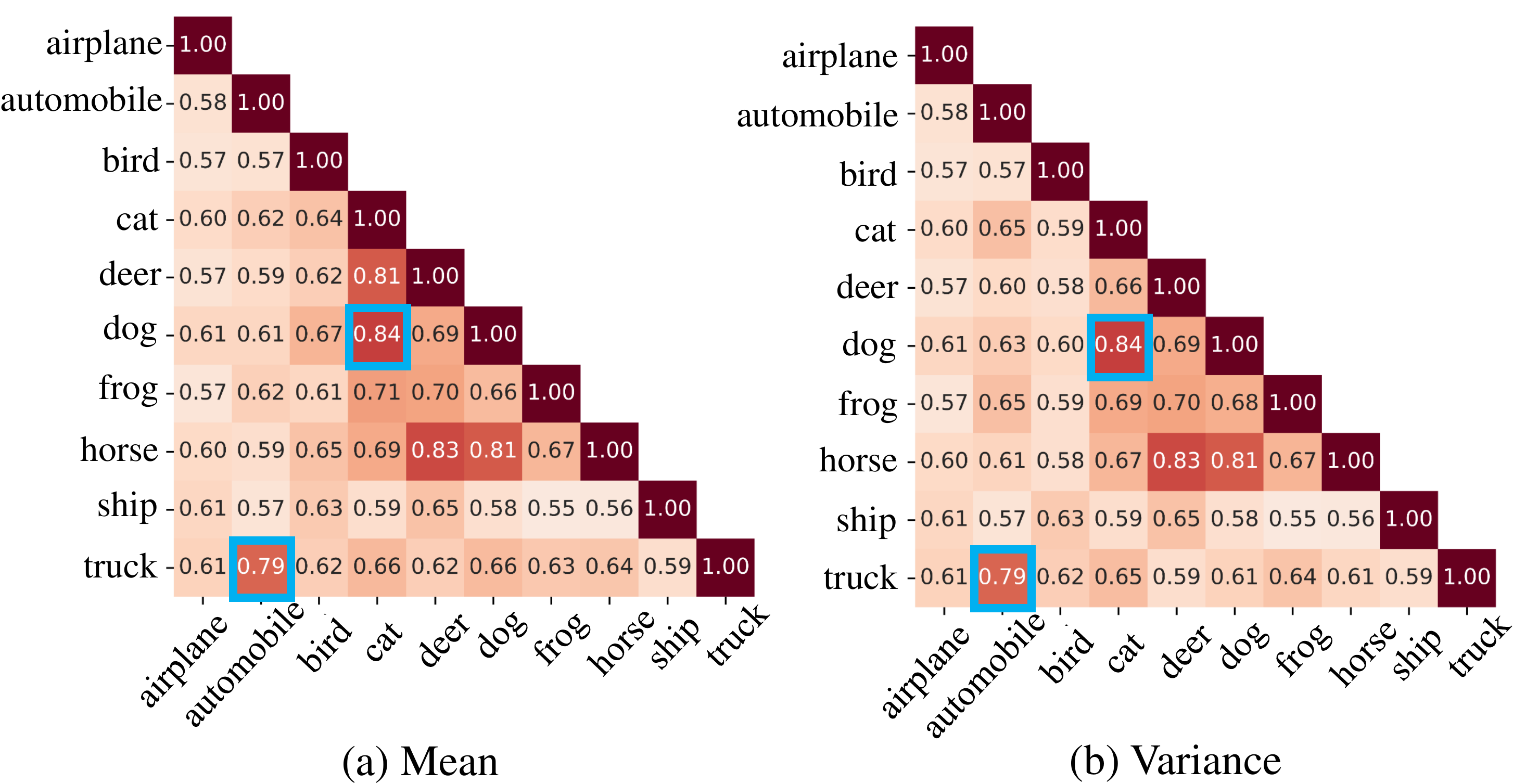}
    \caption{Visualization for the cosine similarities between the means (a) or variances (b) of classes in the feature space of CIFAR10-LT. The feature vectors are computed by the ResNet-32 backbone. The first four classes including $airplane$, $automobile$, $bird$, and $cat$ are head classes, while the others are tail classes. The blue rectangles highlight the high similarities between some head classes and tail classes, such as $cat$ and $dog$, and $automobile$ and $truck$.} 
    \label{fig:similar}
\end{figure}
\begin{figure*}[ht]
    \centering
    \includegraphics[width=0.9\textwidth]{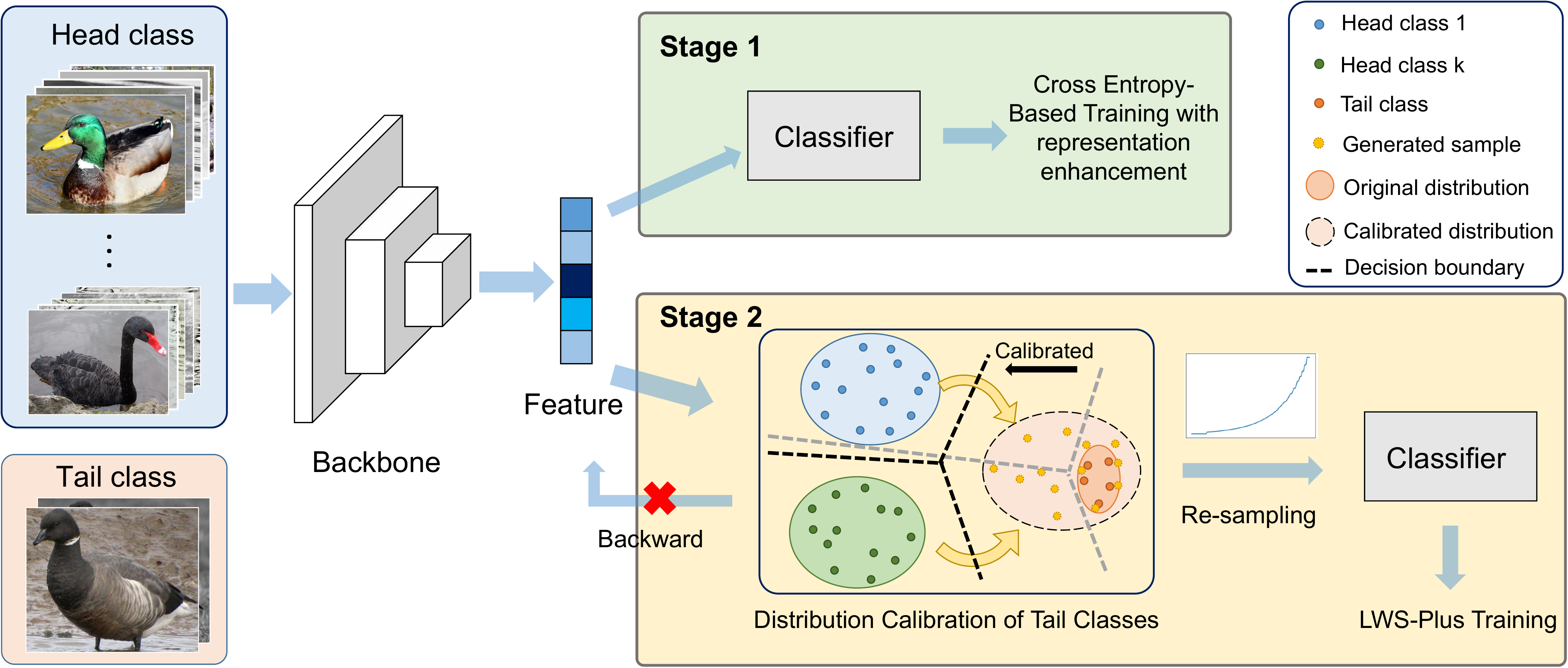}
    \caption{Overview of our proposed LADC method. In Stage 1, we train the backbone with Representation Enhancement. In Stage 2, we freeze the backbone and modify the classifier by re-sampling with distribution calibration and LWS-Plus training.}
    \label{fig:framework}
\end{figure*}
Classification tasks such as image classification and natural language classification are fundamental and essential for evaluating the learning ability of neural networks. Generally, the neural networks are trained and evaluated on balanced datasets including CIFAR and ImageNet~\citep{deng2009imagenet}. However, real-word datasets are imbalanced and usually present long-tailed distributions, that is, head classes contain more training instances while tail classes are with significantly fewer instances. Models trained on long-tailed data will result in obvious performance degradation, e.g., much worse performance for the tail classes compared with the performance for the head classes. The main challenge lies in the sparsity of tail classes, leading to estimation of the decision boundaries severely biased to head classes.

Re-weighting methods~\cite{cui2019class, jamal2020rethinking, cao2019learning} are one popular solution to tackle the challenge. For example, the work~\cite{cui2019class} re-design training loss functions to highlight the contribution of tail classes. 
Recently, \citet{cao2019learning} find that traditional end-to-end re-weighting methods present non-promising results. So they propose to train the model in a normal way in the first stage and tune the model with deferred re-sampling (DRS) and deferred re-weighting (DRW) after annealing the learning rate.
\citet{kang2019decoupling} follow the two-stage training process, and decouple the learning of backbone and classifier to mitigate the impact of imbalanced data on biasing the classifier. Specifically, for re-balancing the decision boundary, they re-train the classifier (cRT) and learn scaling weights for classifier (LWS) via class-balanced sampling in the second training stage.
\citet{zhong2021improving} later propose label-aware smoothing to regularize the classifier. Although the prior studies make some achievements, they have not exploited the inter-class similarity, especially the transferable knowledge from head classes to tail classes.


There are also some studies~\cite{chawla2002smote,li2021metasaug,kim2020m2m} focusing on synthesizing new training instances for tail classes to solve the data sparsity issue. For example, M2m \cite{kim2020m2m} generates minority samples by translating majority samples. MetaSAug \cite{li2021metasaug} produces new training instances via implicit semantic data augmentation \cite{wang2019implicit}. However, these studies pay more attention to the surface features, without explicitly exploring the in-depth feature space.

Inspired by the distribution calibration method (DC) in the few-shot learning~\citep{yang2021free}, which estimates the in-depth feature distribution of support set by utilizing the knowledge from base set, we propose a novel label-aware distribution calibration method (\tool) for our long-tailed scenarios. \tool assumes that the distribution of tail classes can be enriched by similar head classes in the feature space. Following the DC method, \tool also assumes that every dimension in the feature space follows a Gaussian distribution, and similar classes have similar means and variances of the feature representations.  
The proposal of LADC is based on the observation that there exists some similarity between head classes and tail classes. As shown in Fig.~\ref{fig:similar}, we can observe that the statistics of the head class $automobile$ are highly relevant to those of the tail class $truck$. So the feature distribution of the $truck$ class can be better calibrated by virtue of the $automobile$ class which contains sufficient instances for training. 


Specifically, \tool decouples the training process into two stages. In the first training stage, the backbone is trained with data augmentation techniques including Re-balanced Mixup \citep{chou2020remix} and RandAugment \citep{cubuk2020randaugment} to enhance the representation learning. Based on the trained backbone, \tool computes and records the mean feature and co-variance matrix of each head class. In the second training stage, \tool selects the most similar $m$ head classes for computing new 
mean and co-variance of calibrated distribution for every instance in tail classes. Finally, new training instances sampled from the calibrated distributions are employed to rectify the classifier in the first stage.
In addition, we propose LWS-Plus to further improve the classifier rectification ability. In summary, our contributions are as follow:
 \begin{enumerate}
      \item We are the first to exploit the transferable knowledge from head classes to tail classes for estimating the feature distributions of tail classes.
      
      \item We propose a novel method \tool for calibrating the distribution of tail classes by borrowing knowledge from head classes. A new sampling strategy is also proposed based on the calibrated distribution to better balance the classifier.
      
     \item Extensive experiments demonstrate that \approach can significantly outperform state-of-the-art approaches on long-tailed image datasets including CIFAR10-LT, CIFAR100-LT, ImageNet-LT and iNaturalist. Experimentation on long-tailed text classification task indicates that \approach is generalizable to other long-tailed modals.
 \end{enumerate}

\section{Methodology}

In this section, we elaborate on the proposed method \tool. \tool follows a two-stage training framework, in which a backbone is first trained on the original long-tailed dataset and fixed on the second stage. Fig.~\ref{fig:framework} depicts the overview of the proposed \tool.
Equipped with the backbone and the associated representation space, we illustrate how \tool estimates the distribution of tail classes by utilizing the statistics of relevant head classes. Then we introduce the sampling strategy based on the calibrated distribution. We finally describe how \tool enhances the training of backbone and classifiers.

\subsection{Problem Definition}
We follow a typical long-tailed classification setting. Given an imbalanced dataset
  $D = \{\bm{x}_i,y_i\}^N_{i=1}$ with $N$ training instances, where $\bm{x}_i$ denotes a instance and $y_i$ denotes its label.
Assume that the classes are ordered by cardinality, i.e., when class index $p<q$, then $n_p \geq n_q$, where $n_p$ is the number of training instances of the class $p$.
Let $\mathcal{F}$ be the backbone and $\Phi$ be the classifier. Let $\bm{z}^c$ represent the class embedding of class $c$ in the feature space defined by $\mathcal{F}$. Furthermore, we split classes into two groups with head classes denoted as $C_h$ and tail classes as $C_t$. Note that we simplify the notations by dropping class index wherever possible.

\subsection{Label-Aware Distribution Calibration}


The DC method~\citep{yang2021free} for few shot learning focuses on $N$-way $K$-shot task where $N$ is fixed. Since long-tailed scenarios require models to perform well on both head and tail classes, directly adopting DC is non-optimal.
To improve the prediction accuracy of tail classes and preserve the performance on head classes, we propose a novel label-aware distribution calibration approach. 

\subsubsection{Calibrating distribution of tail classes}

Given the trained backbone, \tool calibrates the distributions of tail classes with the help of relevant head classes. 
We assume that each class is Gaussian distributed and head classes are well represented with sufficient data for training. So the distribution of each head class can be approximated by calculating the mean and co-variance of the associated training data:
\begin{equation}
    p(\bm{z}) = N(\bm{z}|\bm{\mu},\bm{\Sigma})
\end{equation}
\begin{equation}
    \bm{\mu} = \frac{\sum_{i=1}^{n} \mathcal{F}(\bm{x}_i)}{n}
\end{equation}
\begin{equation}
    \bm{\Sigma} = \frac{1}{n-1}\sum_{i=1}^{n}(\mathcal{F}(\bm{x}_i)-\bm{\mu})(\mathcal{F}(\bm{x}_i)-\bm{\mu})^T
\end{equation} 
where $\bm{\mu}$ and $\bm{\Sigma}$ denote the mean and variance of the class, and $n$ denotes the number of training instances in the head class.

To more accurately estimate the distribution of each tail class, we employ the statistics of a set of similar head classes $S$ as the prior and compute the posterior by using the feature $\bm{\hat{\mu}}$ of the tail class. The similar head classes $S$ is measured by the Euclidean distance in the feature space:

\begin{equation}
    Set^d = \{-||\bm{\mu}^i-\hat{\bm{\mu}}||^2|i\in C_h\} \\
\end{equation}
\begin{equation}
    S = \{i|-||\bm{\mu}^i-\hat{\bm{\mu}}||^2 \in topm(Set^d)\}
\end{equation}

Equipped with $S$, We can compute the prior distribution of $\bm{\hat{\mu}}$:
\begin{equation}
    \bm{w}_i = \frac{n_i||\bm{\mu}^i-\hat{\bm{\mu}}||^2}{\sum_{j\in \mathbb{S}}n_j||\bm{\mu}^j-\hat{\bm{\mu}}||^2}
\end{equation}

\begin{equation}
    \bm{\mu}_{0}=\sum_{i \in S}\bm{w}_{i}\bm{\mu}^{i} \quad
    \bm{\Sigma}_{0}=\sum_{i \in S}(\bm{w}_{i})^2\bm{\Sigma}^{i} + \alpha \bm{I}
\end{equation}

\begin{equation}
    p(\bm{z}) = N(\bm{z}|\bm{\mu}_{0}, \bm{\Sigma}_{0})
\end{equation}
where $\alpha$ is a hyper-parameter to control the degree of dispersion and $\bm{w}$ is the weight vector which is designed to impose tail classes to learn more from head classes that are more abundant and more similar. 

For each tail class, $\mathcal{F}(\bm{x})$ conditioned on $\bm{z}$ is assumed to follow a Gaussian distribution
\begin{equation}
    p(\mathcal{F}(\bm{x})|\bm{z}) = N(\mathcal{F}(\bm{x})|\bm{z}, \bm{L})
\end{equation} 
where $\bm{L}$ is the co-variance matrix and it is related to the uncertainty of the trained backbone.

Based on the prior distribution, the posterior (calibrated) distribution of each instance in the tail class can be given by
\begin{equation}
    p(\bm{z}|\{\mathcal{F}(\bm{x_i})\}_i) = N(\bm{z}|\bm{\mu}', \bm{\Sigma}')
\end{equation}  By defining $\bm{L}=\beta \bm{\Sigma}_0$, 
we compute $\bm{\mu}'$ and $\bm{\Sigma}'$ as follows:

\begin{equation}\label{equ:BDC_mu}
    \bm{\mu}' = \frac{\beta}{n_s+\beta}\bm{\mu}_{0} + \frac{n_s}{n_s+\beta}\hat{\bm{\mu}}
\end{equation}

\begin{equation}
    \bm{\Sigma}' = \frac{\beta}{n_s+\beta}\bm{\Sigma}_{0}
\end{equation}
where $n_s$ means the sample size for computing $\hat{\bm{\mu}}$ and is set as 1. $\beta$ is a hyper-parameter which can be interpreted as the relative uncertainty of the sample to the prior $\bm{\mu}_0$. Higher $\beta$ indicates a higher confidence of the prior.

\subsubsection{Sampling strategy} In the second training stage, to balance the long-tailed distribution, we define the sampling probability of each class to be:
\begin{equation}
    P_i = \frac{n_1 / n_i^\tau}{\sum_{j}n_1 / n_j^\tau}
\end{equation}
where $n_1$ refers to the number of training instances of the most represented class. $\tau$ is a temperature parameter and it controls the balanced degree of the sampled distribution. Note that we re-sampling head classes from the original distributions; while for the tail classes, we draw samples from the calibrated distributions.

\begin{table*}[ht]
    \centering
    \caption{Overview of datasets used in our experiments. The ``IF'' denotes imbalance factor.}
    \scalebox{0.94}{\begin{tabular}{c|c|c|c|c|c|c}
    \toprule
       Datasets  & \#Class & IF & \#Train Set & Min. Class Size & Max. Class Size & \#Test Set\\ \hline
    CIFAR10-LT & 10 & 10$\sim$200 & 50,000$\sim$11,203 & 500$\sim$25 & 5,000 & 10,000 \\ \hline
    CIFAR100-LT & 100 & 10$\sim$200 & 50,000$\sim$9,502 & 500$\sim$2 & 5,00 & 10,000 \\ \hline
    ImageNet-LT & 1000 & 256 & 115,846 & 5 & 1280 & 50,000 \\ \hline
    iNaturalist2018 & 8142 & 500 & 437,513 & 2 & 1,000 & 24,426 \\ \hline
    THUNEWS-LT & 10 & 100 & 44,723 & 180 & 18,000 & 10,000\\
    \bottomrule
    \end{tabular}
    }
    \label{tab:dataset}
\end{table*}

\subsubsection{Re-balancing classifier} 

In Stage 2, \citet{kang2019decoupling} introduce two methods for classifier adjustment: cRT and LWS. cRT re-trains the classifiers completely, while LWS retains the direction of the classifiers and only adjusts the scales. In the work, we choose LWS due to its superior performance in our experiments.
On the basis of LWS and inspired by logits adjustment work \cite{menon2020long, tang2020long}, we add a learnable bias term to the logits, named LWS-Plus, for adjusting the classification boundaries more flexibly:
\begin{equation}\label{equ:lwsp}
    \hat{\Phi}_i(\bm{x}) = \bm{f}_i \cdot \Phi_i(\bm{x}) + \bm{g}_i 
\end{equation}
where $\Phi_i(\bm{x})$ refers to the logit score of the class $i$, $\bm{f}_i$ denotes the scaling factor of the classifier magnitude, and $\bm{g}_i$ indicates the added bias.

\begin{table*}[ht]
    \centering
    \caption{Top-1 accuracy (\%) on CIFAR10-LT and CIFAR100-LT. Best results are marked in bold.}
    \begin{tabular}{l|r|r|r|r|r|r|r|r}
    \toprule
   Approach & \multicolumn{4}{c|}{CIFAR10-LT} & \multicolumn{4}{c}{CIFAR100-LT} \\ 
    \midrule
    IF & 200 & 100 & 50 & 10 & 200 & 100 & 50 & 10 \\
    \midrule
    Cross-entropy training & 65.87 & 70.14 & 74.94 & 86.18 & 34.70 & 37.92 & 44.02 & 55.73 \\
    Class Balance Loss & 68.77 & 72.68 & 78.13 & 86.90 & 35.56 & 38.77 & 44.79 & 57.57 \\
    Focal Loss & 65.29 & 70.38 & 76.71 & 86.68 & 35.62 & 38.41 & 44.32 & 55.78 \\
    CB Focal Loss & 68.15 & 74.57 & 79.22 & 87.48 & 36.23 & 39.60 & 45.21 & 57.99 \\
    LDAM Loss  & 66.75 & 73.55 & 78.83 & 87.32 & 36.53 & 40.60 & 46.16 & 57.29 \\
    LDAM-DRW  & 74.74 & 77.03 & 81.03 & 88.16 & 38.45 & 42.04 & 46.62 & 58.71 \\
    MCW with LDAM loss& 77.23 & 80.00 & 82.23 & 87.40 & 39.53 & 44.08 & 49.16 & 58.00 \\
    Decoufound-TDE & - & 80.60 & 83.60 & 88.50  & - & 44.10 & 50.30 & 59.60 \\\hline

    cRT & 69.48&73.02 & 79.56& 87.90 & 38.16& 43.30 & 47.37& 57.86\\
    LWS & 67.90&72.44 & 77.77& 87.41 & 37.52& 42.97 & 47.40& 58.08\\
    MiSLAS & 76.73 & 82.10 & 85.70 & 90.00 & 43.53 & 47.00 & 52.30 & 63.20 \\ \hline
    BBN & - & 79.82 & 81.18 & 88.32 & - & 42.56 & 47.02 & 59.12 \\
    LDAM-DRW + \textit{SSP} & - & 77.83 & 82.13 & 88.53 & - & 43.43 & 47.11 & 58.91 \\
    Bag of tricks & - & 80.03 & 83.59 & - & - & 47.83 & 51.69 & -  \\
    
    \hline
    SMOTE & - &71.50 & - & 85.70 & - &34.00 & - & 53.80 \\
    Major-to-Minor & -& 79.10 & -& 87.50 & - & 43.50 & - & 57.60 \\
    MetaSAug-LDAM & 77.35 & 80.66 & 84.34 & 89.68 & 43.09 & 48.08 & 52.27 & 61.28 \\
    
    \hline
     \approach & \textbf{81.56} & \textbf{84.65} & \textbf{87.09} & \textbf{90.81} & \textbf{46.62} & \textbf{50.77} & \textbf{54.94} & \textbf{64.66} \\
    \bottomrule

    \end{tabular}
    
    \label{tab:cifar10}
\end{table*}

\subsubsection{Representation Enhancement}
In Stage 1, Mixup has proven to be an effective regularization strategy for training deep neural networks \citep{zhong2021improving, zhang2021bag}. More recently, Remix has been proposed for imbalanced data \citep{chou2020remix}. In the work, we adopt Remix in the first training stage due to its better performance in the experiments. In addition, we combine RandAugment \citep{cubuk2020randaugment}, a two-stage augmentation policy that uses random parameters in place of parameters tuned by AutoAugment. 
\section{Experiments}

\subsection{Experimental Setup}
\subsubsection{Datasets}
We evaluate \tool on several popular long-tailed classification tasks, including both image and text data. Image datasets include CIFAR-10-LT, CIFAR-100-LT \cite{cui2019class}, ImageNet-LT \cite{cao2019learning} and iNaturalist \cite{van2018inaturalist}. The text dataset includes THUNEWS-LT \cite{sun2016thuctc}. Table \ref{tab:dataset} shows the detailed statistics of the datasets, where the ``IF'' indicates the imbalance factor, i.e., the ratio of the number of the most-represented class to the number of the least-represented class. 

\subsubsection{Baselines}\label{sec:baseline}
We choose several types of comparison methods: (1) one-stage training methods. They include Class-Balanced loss~\citep{cui2019class}, Focal loss~\citep{lin2017focal}, CB-Focal loss~\citep{cui2019class}, LDAM loss~\cite{cao2019learning}, Balanced Softmax~\citep{ren2020balanced}, Meta-class-weight LDAM~\citep{jamal2020rethinking} and a causal model Decoufound-TDE~\citep{tang2020long}. For the LDAM loss, we also consider the results with DRW strategy, i.e., LDAM-DRW. (2) two-stage training methods. The cRT, LWS~\citep{kang2019decoupling} and MiSLAS~\citep{zhong2021improving} reuse the original training data. (3) generative approaches. Feature space augmentation (FSA)~\cite{chu2020feature}, Major-to-Minor~\citep{kim2020m2m} and MetaSAug-LDAM~\citep{li2021metasaug} synthesize new training data. We also involve SMOTE \citep{chawla2002smote}, a widely-used generative method for mitigating oversampling issue. (4) other baselines, including OLTR\cite{liu2019large}, BBN \citep{zhou2020bbn}, SSP \citep{yang2020rethinking}, Bag of tricks~\cite{zhang2021bag} and Hybrid-PSC \cite{wang2021contrastive}.






\subsubsection{Implementation Details.} 
In the experiments, we regard the most frequent classes occupying at least 60\% of the total training instances as head classes, and the remaining classes as tail classes. The number of head classes $m$ selected for computing $S$ is chosen from $\{2, 3\}$. $\alpha$,  $\beta$ and $\tau$ are selected from $0.1\sim 0.2$, $0.4\sim 1.3$ and $1.2\sim 1.3$, respectively. Detailed parameter analysis can be referred to the Appendix.

For CIFAR10-LT and CIFAR100-LT, we use ResNet-32 \citep{he2016deep} as backbone following the work \cite{cao2019learning}. We first train the backbone with representation enhancement for 400 epochs with five warm up steps. The base learning rate is 0.1 and we conduct a multi-step learning rate schedule which decreases learning rate by 0.01 at the $320^{th}$ and $360^{th}$ epochs. In the second training stage, we freeze the backbone and train classifier with \approach for 30 epochs during which the learning rate drops by 0.1 at the $10^{th}$ and $20^{th}$ epochs. 

For ImageNet-LT dataset, we choose ResNet-10 and ResNet-50 as the backbone and train for 200 epochs in the first stage. The learning rate is initialized as 0.2 and decreases by 0.1 at the $120^{th}$ and $160^{th}$ epochs. For the iNaturalist dataset, we train our approach for 100 epochs employing the cosine learning rate schedule~\cite{loshchilov2016sgdr} for 100 epochs with ResNet-50 in the first stage. The training strategy in the second stage is similar as that in training the CIFAR datasets.


We use SGD optimizer with the momentum at 0.9 and weight decay at $5\cdot10^{-4}$ for all the experiments. The experimentation is run on a single NVIDIA Tesla V100 with 32GB graphic memory.


\subsection{Experiment Results}
\subsubsection{Main results on CIFAR}
Table \ref{tab:cifar10} shows the results on CIFAR10-LT and CIFAR100-LT. As can be seen, the proposed LADC consistently outperforms all the baselines. For the re-weighting approaches (i.e., the top 7 baselines), \approach increases the performance by at least 4.31\% and 6.54\% on average for CIFAR10-LT and CIFAR100-LT, respectively. Comparing with the best two-stage approach MiSLAS, \approach shows an improvement by 2.72\% on CIFAR100-LT. \approach also performs better than the state-of-the-art generative approach MetaSAug, presenting an increase by 3.02\% and 3.05\% on average for the two datasets, respectively. The results indicate the effectiveness of the proposed approach in the long-tailed image classification. 


\subsubsection{Main results on ImageNet-LT} We conduct experiments on two backbones, i.e., ResNet-10 and ResNet-50. As shown in Table~\ref{tab:cifar100-50}, \approach achieves the best performance among all the approaches for different backbones. Specifically, \approach outperforms the baselines by at least 0.47\% and 1.13\% for the two backbones, respectively. The results imply that representations learned by a larger model could better benefit \approach.

\begin{table}[t]
\centering
\caption{\label{tab:cifar100-50}Top-1 accuracy (\%) on ImageNet-LT. Best results are marked in bold. BALAMs denotes Balanced Softmax loss with meta sampler.}
\begin{tabular}{l|rr}
\toprule

Approach & ResNet-10 & ResNet-50\\ \midrule
Focal Loss & 30.50 & 43.70 \\
LDAM-DRW  & 40.73 & 48.80\\
BALAMs & 41.80 & 50.04\\
OLTR & 35.60 & 41.90\\
Remix &  37.58 & 46.19 \\ 
cRT & 41.80 & 47.30 \\ 
LWS & 41.40 & 47.70\\ 
FSA & 35.20 & -\\
cRT+\textit{SSP}  & 43.20 & 51.30 \\
Bag of Tricks & 43.13 & - \\
MiSLAS & - & 51.47 \\
MetaSAug  &- & 50.52 \\\hline
\approach & \textbf{43.67} & \textbf{52.60} \\\bottomrule

\end{tabular}

\end{table}

\subsubsection{Main Results on iNaturalist}
Table \ref{tab:inat} presents the results on large real-world dataset iNaturalist with ResNet-50. We can observe that \approach increases the accuracy of baseline approaches by 0.58\%$\sim$8.21\%, which shows the capability of \tool in handling large long-tailed datasets.


\subsubsection{Main results on the text dataset THUNEWS-LT}
We choose a commonly-used subset of Chinese news dataset \textit{THUCNews} \citep{sun2016thuctc} which contains ten classes\footnote{https://github.com/649453932/Chinese-Text-Classification-Pytorch}. We build its long-tailed version with the imbalance factor at 100 and use the pre-trained embedding model provided by \citet{li2018analogical}. We choose three models including TextCNN~\cite{kalchbrenner2014convolutional}, TextRNN \citep{liu2016recurrent} and Transformer \citep{vaswani2017attention} as the backbone. We remove the representation enhancement component due to the inapplicability of the data augmentation techniques for the text data. The results illustrated in Table~\ref{tab:nlp} show that \approach achieves the highest accuracy, increasing the best baselines by 0.5\%, 1.49\% and 0.99\% on three models, respectively. The results demonstrate the generalizability and effectiveness of \approach in other long-tailed modals.
\begin{table}[t]
    \centering
    \caption{Top-1 accuracy (\%) on iNaturalist2018. All the models are trained up to 100 epochs.}
    \label{tab:inat}
    \begin{tabular}{l|r}
    \toprule
     Approach  & ResNet-50 \\
     \midrule
      CE & 57.16   \\
      CB-Focal & 61.12  \\
      LDAM-DRW & 68.00  \\
      BBN & 66.29 \\
      Decoufound-TDE & 65.20 \\
      cRT+\textit{SSP} & 68.10 \\
      Hybrid-PSC & 68.10  \\
      MetaSAug & 68.75 \\
    \hline
      \approach & \textbf{69.33} \\
      \bottomrule
      
    \end{tabular}
\end{table}

\begin{table}[ht]
    \centering
    \caption{Top-1 accuracy (\%) on Chinese text classification. CB loss and BSCE denote Class Balanced loss and Balanced Softmax, respectively.}
    \begin{tabular}{c|rrr}
    \toprule
       Approach  & TextCNN & TextRNN & Transformer \\
       \midrule
        CE   & 78.65 & 78.79 & 79.12 \\
        CB Loss & 80.44 & 80.83 & 80.43 \\
        LWS & 83.97 & 81.45 & 82.88 \\
        BSCE & 84.28 & 81.06 & 83.02 \\
        \midrule
        \approach & \textbf{84.78} & \textbf{82.94} & \textbf{84.01} \\
        \bottomrule
        
    \end{tabular}
    
    \label{tab:nlp}
\end{table}

\begin{table}[ht]
    \centering
    \caption{Ablation study on CIFAR-100-LT. RE: the representation enhancement module in Stage 1. Sampling: sampling from the distributions calibrated by \approach in Stage 2.}
    \label{tab:module}
    \scalebox{0.9}{\begin{tabular}{cccc|ccc}
    \toprule
    \multicolumn{4}{c}{Module} & \multicolumn{3}{c}{CIFAR100-LT} \\
    \midrule
     RE & Sampling & LWS&LWS-Plus & 200 & 100 & 50 \\
    \XSolidBrush & \XSolidBrush & \XSolidBrush & \XSolidBrush & 34.70 & 37.92 & 44.02 \\
     \Checkmark & \XSolidBrush& \XSolidBrush&  \XSolidBrush& 38.12& 42.30 & 46.72\\ 
     \Checkmark & \Checkmark& \Checkmark & \XSolidBrush& 45.29& 49.22 & 54.00\\ 
      \Checkmark & \Checkmark & \XSolidBrush & \Checkmark &\textbf{46.62} & \textbf{50.77} & \textbf{54.94}\\ 
      \bottomrule
    \end{tabular}
    }
\end{table}

\subsection{Discussion}

\subsubsection{Ablation Study} We perform an ablation study to analyze the impact of each module in \approach. We follow the same two-stage training implementation throughout this study, and the results are shown in Table \ref{tab:module}. Comparing the first two rows of the table, we can see that RE benefits the classification performance, as expected. In addition, we can observe that sampling from the calibrated distribution brings further improvement, as shown by the third row. Comparing the last two rows, we can achieve that the proposed LWS-Plus better rectifies the decision boundary than LWS.



\subsubsection{Results Analysis}
Following \citet{wang2020long}, we divide the classes into three groups according to the number of training instances associated with each class, i.e., Many-shot ($>$100 instances), Medium-shot ($100\sim 20$ instances) and Few-shot ($<$20 instances). The results of different groups are shown in Table \ref{tab:freqency}. Comparing with the model trained by Cross-Entropy (CE), the baselines mainly focus on boosting the performance of the few-shot group with slight improvement on the medium-shot group. Besides, all the baselines expect for Remix show a performance degradation on the many-shot group. In contrast, the proposed \approach significantly enhances the accuracy on the few-shot and medium-shot groups, while achieving almost similar performance on the many-shot group as CE. The results indicate the effectiveness of \tool on both head and tail classes.


\begin{table}[ht]
    \centering
    \caption{Top-1 accuracy (\%) of different groups on CIFAR100-100.}
    \label{tab:freqency}
    \begin{tabular}{l|r|rrr}
    \toprule
      Approach   & All & Many & Medium & Few \\
      \midrule
       CE  & 37.9 & 65.2 & 36.3 & 8.0 \\
       \hline
       Remix & 39.7 & \textbf{69.4} & 37.9 & 7.2 \\
       OLTR & 41.2 & 61.8 &  41.4 & 17.6\\
       LDAM-DRW & 42.0 & 61.5 & 41.7 & 20.2 \\
       $\tau$-norm & 41.4 & 58.8 & 38.1 & 24.7 \\
       cRT & 43.3 & 64.0 & 44.8 & 18.1 \\
       \hline
       \approach & \textbf{50.8} & 64.5 & \textbf{52.6} & \textbf{32.5} \\
       \bottomrule
    \end{tabular}
\end{table}

\subsubsection{Calibration strategy} 
To analyze the benefit of our distribution calibration strategy, we compare with some alternative variants of our method. One variant is LADC\textsubscript{Average}, indicating calibrating distribution for the whole tail class, i.e., $\bm{\hat{\mu}}$ in Equ. (\ref{equ:BDC_mu}) is the mean feature of the tail class. In addition, we compare with the DC method in the few-shot learning field \cite{yang2021free}. Note that the sampling strategy is the same for all the comparisons. The results are listed in Table \ref{tab:aba1}. As can be seen, our proposed \approach significantly outperforms the DC method, which demonstrates that \tool is more effective than DC in the long-tailed scenario. In addition, comparing \approach with LADC\textsubscript{Average}, we can achieve that instance-based distribution calibration can better estimate the distributions of classes than class-based calibration. 

\subsubsection{Comparison with different classifier re-balancing approaches in Stage 2.} Table \ref{tab:abla2} shows the results of different classifier re-balancing approaches in the second training stage. The experiments are also conducted on CIFAR100-LT. We choose class-balanced sampling (CBS), the state-of-the-art regularization method - label-aware smoothing (LAS) and our proposed label-aware distribution calibration combined with cRT and LWS, respectively, to re-balance the classifier in Stage 2. We can observe that \approach significantly improves the performance of class-balanced sampling method (CBS). Besides, \approach outperforms LAS under different training approaches in Stage 2. The results show the effectiveness of the proposed distribution calibration method. 



\subsubsection{Visualization of Generated Samples}
\begin{figure*}[t]
    \centering
    \includegraphics[width=0.9 \textwidth]{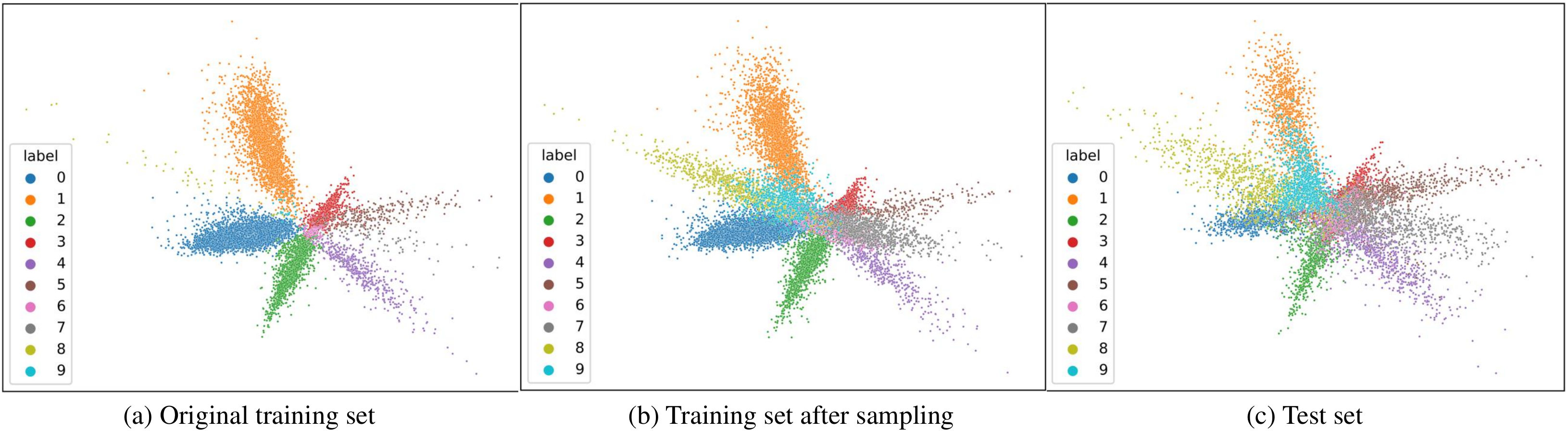}
    \caption{Visualization of distribution estimation on CIFAR10-200 with the ResNet-32 backbone in the feature space. 
    }
    \label{fig:visual}
\end{figure*}
Fig. \ref{fig:visual} presents the visualization of feature distribution of training set and test set of CIFAR10-200. Visualization is conducted by adding a fully-connected layer ($D*2$ where $D$ is the dimension of features) which projects feature into 2-dimension space following \citet{DBLP:conf/icml/LiuWYY16}
Fig. \ref{fig:visual} (a) shows the original training set, where we can see the tail classes (e.g., label 7,8,9) have few training instances and their distributions are significantly different from those in the test set, as illustrated in Fig. \ref{fig:visual} (c). 

After sampling via \approach, we can observe in Fig. \ref{fig:visual} (b) that the distributions of tail classes are obviously extended and more closer to the test set. This indicates that the difference between the distributions of the original training set and test set is relieved by sampled instances. Meanwhile, \approach can well preserve the distributions of head classes. 

\begin{table}[t]
    \centering
    \caption{Study on different distribution calibration strategies.}
    \begin{tabular}{l|rrrr}
    \toprule
       \multirow{2}{*}{Approach}  &  \multicolumn{4}{c}{CIFAR100-LT}\\
          & 200 & 100 & 50 & 10\\
    \midrule
    DC & 44.10& 48.35& 51.69& 62.59\\
    LADC\textsubscript{Average} & 41.74 & 48.01& 47.25& 62.34\\
    \approach(ours) & \textbf{46.62} & \textbf{50.77} & \textbf{54.94} & \textbf{64.66} \\
    \bottomrule
          
    \end{tabular}
    
    \label{tab:aba1}
\end{table}

\begin{table}[]
    \centering
    \caption{Comparison on different classifier modification approach in Stage 2 given a backbone trained with representation enhancement module. CBS and LAS denote class-balanced sampling and label-aware smoothing, respectively.}
    \scalebox{0.92}{\begin{tabular}{cc|rrr}
    \toprule
       \multicolumn{2}{c}{Approach}  & \multicolumn{3}{c}{CIFAR100-LT} \\
       \midrule
       \multicolumn{2}{c|}{Imbalance Factor}   & 200 & 100 & 50 \\
         \midrule
         \multirow{3}{*}{cRT}& CBS & 45.20 & 50.14 & 54.18\\
          & LAS & 45.87 & 50.02 & 53.10 \\
          & \approach & 46.92 & 50.13 & 55.24\\ 
          \midrule
          \multirow{3}{*}{LWS}& CBS & 43.77 & 48.23 & 53.04 \\
          & LAS & 44.02 & 48.21 & 52.59 \\
         & \approach &  45.29& 49.22 & 54.00\\ 
     
      \bottomrule
    \end{tabular}
    }
    \label{tab:abla2}
\end{table}
          

\section{Related work}

\textbf{Re-sampling and Re-weighting.} 
Re-sampling incorporates two approach types, i.e., over sampling tail classes \citep{shen2016relay, buda2018systematic, byrd2019effect} and under sampling head classes \citep{buda2018systematic, japkowicz2002class}. Over sampling tail classes can augment training data of tail classes, which may cause severe over fitting problems on tail classes. Under sampling head classes is feasible to prevent head classes from dominating training process, however, it inevitably degrades the generalization ability of models for the reduction of head data.

Broadly speaking, a group of works aim to mitigate the issue of long-tail distribution by re-weighting training instances in the objective function \citep{cui2019class, tan2020equalization, lin2017focal, ren2020balanced, cao2019learning, tan2021equalization}. For example, Focal loss \citep{lin2017focal} determines the weights according to the model's confidence of the training instances. Several other works determine the weights in inverse proportion to the class frequency. \citet{cui2019class} calculate the weights according to the effective number of each classes. Inspired by the generalization ability of adding margins,  \citet{cao2019learning} assign a label-aware margin to each class. While the re-weighting methods in \citet{cui2019class} and \citet{cao2019learning} determine class weights in a pre-defined manner, meta-class-weight \cite{jamal2020rethinking} adopts a learnable class weights via meta-learning. Although these re-weighting methods enable the training of neural networks in a end-to-end fashion, they result in sub-optimal performance in comparison with two-stage training methods, potentially due to the distortion in representation learning caused by early re-weighting.

\textbf{Two-stage training.} \citet{cao2019learning} propose a two-stage training methods, in which re-weighting or re-sampling is only introduced in the second stage. Decoupled learning is another form of two-stage training that is introduced by \citet{kang2019decoupling} who claim that training on imbalance dataset biases the classifier instead of representation learning. In the first stage, \citet{kang2019decoupling} train a backbone using a traditional cross-entropy loss. Then the backbone is fixed and the classifier is re-balanced by one of the following methods: classifier re-training (cRT),  classifier normalization ($\tau$ norm) and learnable classifier weight scaling (LWS). MiSLAS by \citet{zhong2021improving} is also a decoupled learning approach, which utilizes label-aware smoothing to regularize the classifier. Although simple and effective, there are some issues not properly addressed by these two-stage methods. For example, class-balance sampling can still cause over-fitting to tail classes during the second training stage. Another issue is the distribution mismatch of tail classes. Our method aims to address these issues via label-aware distribution calibration and generative oversampling. 

\textbf{Data synthesis approach.}
Synthesizing new instances has been used widely to construct a balanced dataset.
SMOTE proposed by \citet{chawla2002smote} generates new data through convex combination of a data point and its neighbors. Based on SMOTE, several variants 
have been introduced \citep{han2005borderline, mullick2019generative}. Although they can potentially alleviate the issue of over-fitting, these methods cannot address the distribution mismatch issue of tail classes. More recently, there have been works proposed to generate data for tail classes relying on head classes. For example, M2m \citep{kim2020m2m} synthesizes data for tail classes via translating samples from head classes. And MetaSAug \citep{li2021metasaug} combines implicit semantic data augmentation and meta-learning to generate new training instances. These generative methods are computationally complicated and still need to combine re-weighting methods.

\section{Conclusion}
In this paper, we propose a novel label-aware distribution calibration (\approach) method for long-tailed classification. \approach can calibrate the distribution of tail classes by borrowing knowledge from well-represented head classes. Based on the decouple learning framework, we also explore representation enhancement and classifier re-balancing technique. Extensive experiments on several long-tailed image and text classification datasets demonstrate the effectiveness of \approach. 

\bibliography{sigproc.bib}

\end{document}